\documentclass{article}

\usepackage[utf8]{inputenc}
\usepackage[T1]{fontenc}
\usepackage[english]{babel}
\usepackage[a4paper, margin=2cm]{geometry}
\usepackage{graphicx}
\usepackage{booktabs}
\usepackage{longtable}
\usepackage{array}
\usepackage{amsmath}
\usepackage{amssymb}
\usepackage{float}
\usepackage{xcolor}
\usepackage{hyperref}
\usepackage{url}
\usepackage{enumitem}
\usepackage{tikz}
\usetikzlibrary{arrows.meta, positioning}

\hypersetup{
  colorlinks=true,
  linkcolor=blue,
  urlcolor=blue,
  citecolor=blue
}

\newcommand{\datasetname}{METBRA25Y}
\newcommand{\datasetversion}{v1.0.0}
\newcommand{\datasetdoi}{10.5281/zenodo.19964979}
\newcommand{\dataseturl}{https://doi.org/\datasetdoi}
\newcommand{\codeurl}{https://doi.org/\datasetdoi}
\newcommand{\licensedata}{Creative Commons Attribution 4.0 International (CC BY 4.0)}
\newcommand{\licensecode}{MIT}
\newcommand{\contactemail}{lima.castro@unesp.br}

\title{\datasetname: Brazil Surface Meteorology Archive with Harmonized Variables and Quality Control}
\author{
  Matheus Lima Castro\textsuperscript{1,2,3}\thanks{Correspondence: \texttt{lima.castro@unesp.br}}
  \and
  William Dantas Vichete\textsuperscript{1}
  \and
  Leopoldo Lusquino Filho\textsuperscript{1,3}
  \\[0.75em]
  \small \textsuperscript{1} São Paulo State University (UNESP), Sorocaba, Brazil.\\
  \small \textsuperscript{2} HairU Serviços e Tecnologia LTDA, Sorocaba, Brazil.\\
  \small \textsuperscript{3} RECOD.ai, University of Campinas (UNICAMP), Campinas, Brazil.
}
\date{}

\begin{document}
\maketitle

\begin{abstract}
This data paper describes \datasetname, a harmonized archive of hourly surface meteorological observations from Brazil derived from public historical records of the Instituto Nacional de Meteorologia (INMET). The dataset was designed to support reproducible environmental, climatological, hydrological, agricultural, urban-risk, and machine-learning studies that require station-level meteorological time series with standardized variable names and explicit quality-control metadata. The processing workflow ingests annual INMET archives, parses station metadata from raw file headers, normalizes heterogeneous Portuguese column names into a canonical schema, constructs hourly timestamps, consolidates observations by city and station, and exports compressed CSV files together with station manifests, per-station quality flags, daily precipitation aggregates, variable-level failure summaries, and missing-data audits. The quality-control protocol follows a two-stage strategy: first, physically implausible values are converted to missing values and flagged; second, temporal and cross-variable consistency checks generate diagnostic flags without necessarily overwriting the original measurements. The resulting package covers observations between 2000 and 2025, with station-specific temporal coverage, and includes key meteorological variables such as precipitation, air temperature, dew point, relative humidity, atmospheric pressure, wind speed, wind gust, wind direction, and global solar radiation. Based on the summary files included in the current release snapshot, the archive contains 616 unique station codes across variable summaries, of which 605 have coordinates within a broad Brazil plausibility envelope. This paper documents the dataset provenance, file organization, harmonized schema, quality-control rules, technical validation outputs, limitations, and recommended usage practices.
\end{abstract}


\section*{Data availability and license}
\textbf{Dataset version:} \datasetversion. \\
\textbf{Dataset DOI:} \href{https://doi.org/\datasetdoi}{\datasetdoi}. \\
\textbf{Dataset landing page:} \href{\dataseturl}{\dataseturl}. \\
\textbf{Code and documentation:} included in the Zenodo release at \href{\codeurl}{\codeurl}. \\
\textbf{Dataset license:} \licensedata. \\
\textbf{Code license:} \licensecode. \\
\textbf{Contact:} \texttt{\contactemail}.

\noindent The primary source records are public meteorological observations made available by INMET through its historical-data services and automatic-station catalog \cite{inmet_bdmep_2026,inmet_catalogo_automaticas_2026}. Version \datasetversion{} of \datasetname{} is released as a frozen Zenodo record with DOI, citation metadata, changelog, and checksums, which improves findability and long-term reuse in line with FAIR-oriented publication practices \cite{wilkinson2016fair}. The Zenodo release includes the processed data snapshot, code, documentation, citation metadata, changelog, and checksum manifest required to identify and reuse version \datasetversion{} \cite{metbra25y_zenodo_2026}. The dataset should be cited as a versioned Zenodo record \cite{metbra25y_zenodo_2026}.

\section{Introduction}

Surface meteorological observations are essential for environmental monitoring, climate-risk assessment, hydrological modeling, agricultural decision support, urban flooding studies, renewable-energy analysis, and data-driven weather applications. In Brazil, INMET maintains a national observing network and provides public access to historical meteorological records, including annual archives for automatic stations \cite{inmet_bdmep_2026,inmet_catalogo_automaticas_2026}. However, the direct reuse of these records in computational pipelines can be hindered by heterogeneous file headers, Portuguese variable names with accent and unit variations, station-specific metadata formats, missing values, sentinel values, and quality issues that are common in long observational archives.

\datasetname{} addresses this barrier by transforming the raw annual INMET files into a harmonized, station-aware, and quality-controlled archive. The dataset is not intended to replace the official INMET source. Instead, it provides a reproducible processing layer that standardizes variable names, consolidates observations, records quality-control flags, and exposes summary products designed for downstream modeling. The release is aligned with FAIR principles for scientific data stewardship \cite{wilkinson2016fair} and with established guidance for climatological practice and meteorological observation management \cite{wmo2011climatological,wmo_no8_2026}.

The contribution of \datasetname{} is fourfold:
\begin{enumerate}[label=(\roman*)]
    \item it provides a canonical variable schema for major hourly surface meteorological variables;
    \item it consolidates raw observations into compressed city-level files with station metadata preserved at row level;
    \item it documents both destructive and non-destructive quality checks through explicit flags; and
    \item it includes summary files that quantify missing or invalid hours by station and variable.
\end{enumerate}

Conceptually, \datasetname{} is closer to curated observational archives and quality-controlled station collections than to a simple file mirror. In that sense, it is informed by prior large-scale surface-station efforts such as the Integrated Surface Database, HadISD, and GHCN-Daily, as well as by more recent dataset papers that emphasize explicit technical validation and versioned public release \cite{smith2011isd,dunn2012hadisd,dunn2016expanding_hadisd,menne2012ghcnd,nishimura2023sigma}. For quality screening, especially in the context of precipitation and operationally noisy station records, the present pipeline also draws on the broader literature on automated quality assurance and rainfall-specific QC workflows \cite{durre2010qa,lewis2021hourly_rain,costa2021gapfill_qc}.

\section{Source Data and Scope}

The raw inputs are annual INMET archives, conventionally named by year, such as \texttt{2000.zip}, \texttt{2001.zip}, ..., \texttt{2025.zip}. Each archive contains station CSV files with meteorological observations and metadata embedded in the file header. The \datasetname{} pipeline was configured for the date interval from \texttt{2000-01-01} to \texttt{2025-12-31}; however, the actual coverage is station-specific because stations enter, leave, or present gaps in the archive over time. In addition, the final year should be interpreted as source-dependent and potentially partial, because station availability and upstream archive completeness vary across the network.

The raw annual INMET archives used to build \datasetname{} \datasetversion{} span \texttt{2000.zip} to \texttt{2025.zip}, with \texttt{2025} corresponding to the latest source archive available at the time of collection. Because the upstream INMET services may be updated over time, \datasetname{} should be interpreted as a frozen derived snapshot of the source files used to generate version \datasetversion{}.

The source files contain hourly observations from automatic meteorological stations. The harmonized output keeps the observations at hourly resolution and builds a canonical \texttt{DATAHORA} timestamp by combining the date and UTC hour fields when a timestamp is not already available. Timestamps are parsed as UTC and then exported without timezone information after normalization, preserving chronological ordering while avoiding timezone ambiguity in most tabular workflows. The resulting archive complements broader Brazilian weather and climate products, including daily gridded datasets derived from national station observations, by preserving the original station-level and hourly structure needed for sub-daily analyses, event detection, and methodological benchmarking \cite{xavier2016brazil_gridded,xavier2022brdwgd}.

The current release snapshot used to draft this paper contains nine variable-level summary files in the \texttt{summaries/} directory. These summaries report one row per station and variable, including the station code, station name, coordinates, start and end years, and the number of post-filter missing or invalid hours. Table~\ref{tab:release_summary} summarizes these files.

\begin{table}[H]
\centering
\caption{Variable-level summary files included in the current \datasetname{} snapshot. \texttt{FAIL\_HOURS} denotes hours that are missing or invalid after the quality-control pipeline for the corresponding variable. Values were computed from the uploaded \texttt{summaries/*\_summary.csv} files.}
\label{tab:release_summary}
\scriptsize
\begin{tabular}{lrrrrrr}
\toprule
Summary file & Station codes & Earliest start & Latest end & Median \texttt{FAIL\_HOURS} & Mean \texttt{FAIL\_HOURS} & P90 \texttt{FAIL\_HOURS}\\
\midrule
\texttt{dewpoint} & 616 & 2000 & 2025 & 15665 & 22275 & 44212\\
\texttt{humidity} & 616 & 2000 & 2025 & 15714 & 22363 & 44263\\
\texttt{precipitation} & 615 & 2000 & 2025 & 17925 & 26006 & 51164\\
\texttt{pressure} & 616 & 2000 & 2025 & 13324 & 20278 & 40666\\
\texttt{solar} & 616 & 2000 & 2025 & 68339 & 68316 & 96884\\
\texttt{temp} & 616 & 2000 & 2025 & 13015 & 20001 & 39290\\
\texttt{wind\_dir} & 615 & 2000 & 2025 & 16571 & 24469 & 48531\\
\texttt{wind\_gust} & 615 & 2000 & 2025 & 16539 & 24281 & 48072\\
\texttt{wind\_speed} & 615 & 2000 & 2025 & 16217 & 24835 & 48207\\
\bottomrule
\end{tabular}
\end{table}

\section{Dataset Construction and Harmonization}

\subsection{Processing workflow}

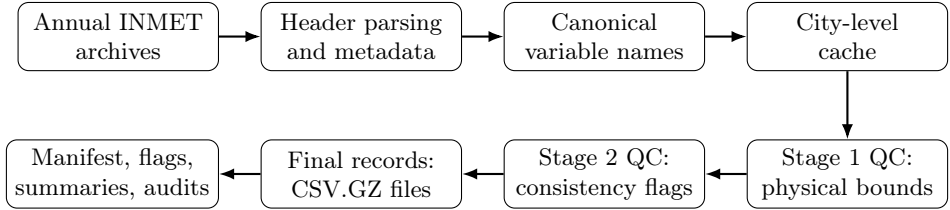
\begin{figure}[H]
\centering
\begin{tikzpicture}[
  node distance=0.9cm and 0.55cm,
  box/.style={rectangle, rounded corners, draw=black, align=center, minimum height=0.9cm, minimum width=2.65cm, font=\small},
  arrow/.style={-{Latex[length=2mm]}, thick}
]
\node[box] (raw) {Annual INMET\\archives};
\node[box, right=of raw] (parse) {Header parsing\\and metadata};
\node[box, right=of parse] (canon) {Canonical\\variable names};
\node[box, right=of canon] (cache) {City-level\\cache};

\node[box, below=of cache] (qc1) {Stage 1 QC:\\physical bounds};
\node[box, left=of qc1] (qc2) {Stage 2 QC:\\consistency flags};
\node[box, left=of qc2] (outputs) {Final records:\\CSV.GZ files};
\node[box, left=of outputs] (summaries) {Manifest, flags,\\summaries, audits};

\draw[arrow] (raw) -- (parse);
\draw[arrow] (parse) -- (canon);
\draw[arrow] (canon) -- (cache);
\draw[arrow] (cache) -- (qc1);
\draw[arrow] (qc1) -- (qc2);
\draw[arrow] (qc2) -- (outputs);
\draw[arrow] (outputs) -- (summaries);
\end{tikzpicture}
\caption{Reproducible construction workflow for \datasetname. The first stage ingests and harmonizes raw INMET files; the second stage consolidates observations, applies quality-control rules, and exports final data records and validation products.}
\label{fig:pipeline}
\end{figure}

Figure~\ref{fig:pipeline} summarizes the \datasetname{} construction workflow. The pipeline is implemented in Python and is distributed in the Zenodo release together with the scripts \texttt{METBRA25Y.py} and \texttt{generatemap.py}, as well as a \texttt{requirements.txt} file. The uploaded dependency specification lists \texttt{pandas==2.2.2}, \texttt{numpy==1.26.4}, \texttt{tqdm==4.66.4}, and \texttt{Unidecode==1.3.8} as core dependencies; optional extensions may include \textit{pyarrow} or other libraries for alternative storage formats. Future releases should additionally record the exact Python interpreter version and, where applicable, a source-code commit hash or release tag for stricter computational reproducibility.

\subsection{Ingestion and station metadata parsing}

For each annual archive, the pipeline extracts raw station CSV files into a temporary directory and reads their header lines using the \texttt{latin-1} encoding. Header parsing is designed to be robust to accent marks, separator variations, and inconsistent naming. The extracted metadata include:
\begin{itemize}
    \item \texttt{STATION\_NAME}: station name;
    \item \texttt{STATION\_CODE}: station or WMO-like station code;
    \item \texttt{UF}: Brazilian state abbreviation;
    \item \texttt{LAT} and \texttt{LON}: station coordinates;
    \item fallback city and state information inferred from file names when needed.
\end{itemize}

These metadata are propagated to the observational rows so that station identity and location remain available after city-level consolidation. The generated manifest is expected to contain one row per station with \texttt{city}, \texttt{uf}, \texttt{STATION\_CODE}, \texttt{STATION\_NAME}, \texttt{LAT}, \texttt{LON}, \texttt{start\_year}, \texttt{end\_year}, \texttt{rows}, and \texttt{file}.

\subsection{Canonical variable schema}

The raw INMET files contain Portuguese column names with spelling, accent, capitalization, and unit variations. \datasetname{} maps these columns to a stable canonical schema. Table~\ref{tab:variables} lists the main variables.

\begin{longtable}{p{0.27\textwidth}p{0.16\textwidth}p{0.47\textwidth}}
\caption{Main canonical variables in \datasetname.}\label{tab:variables}\\
\toprule
Canonical name & Unit & Description\\
\midrule
\endfirsthead
\toprule
Canonical name & Unit & Description\\
\midrule
\endhead
\texttt{DATAHORA} & UTC hour & Normalized hourly timestamp.\\
\texttt{STATION\_CODE} & -- & Station code extracted from header or file name.\\
\texttt{STATION\_NAME} & -- & Station name extracted from the raw INMET header.\\
\texttt{LAT}, \texttt{LON} & degrees & Station latitude and longitude.\\
\texttt{PRECIPITATION\_MM} & mm & Hourly total precipitation.\\
\texttt{PRESSURE\_MB} & hPa / mB & Atmospheric pressure at station level.\\
\texttt{PRESSURE\_MAX\_MB} & hPa / mB & Maximum pressure in the previous hour.\\
\texttt{PRESSURE\_MIN\_MB} & hPa / mB & Minimum pressure in the previous hour.\\
\texttt{SOLAR\_RADIATION\_KJ\_M2} & kJ m$^{-2}$ & Global solar radiation.\\
\texttt{TEMP\_C} & $^\circ$C & Dry-bulb air temperature.\\
\texttt{DEWPOINT\_C} & $^\circ$C & Dew-point temperature.\\
\texttt{TEMP\_MAX\_C} & $^\circ$C & Maximum air temperature in the previous hour.\\
\texttt{TEMP\_MIN\_C} & $^\circ$C & Minimum air temperature in the previous hour.\\
\texttt{RH\_\%} & \% & Relative humidity.\\
\texttt{RH\_MAX\_\%} & \% & Maximum relative humidity in the previous hour.\\
\texttt{RH\_MIN\_\%} & \% & Minimum relative humidity in the previous hour.\\
\texttt{WIND\_DIR\_DEG} & degrees & Wind direction.\\
\texttt{WIND\_GUST\_MPS} & m s$^{-1}$ & Maximum wind gust.\\
\texttt{WIND\_SPEED\_MPS} & m s$^{-1}$ & Hourly wind speed.\\
\bottomrule
\end{longtable}

\subsection{Numeric coercion and sentinel treatment}

After column harmonization, numeric fields are coerced from text to numeric values. Decimal-comma representations are converted where necessary, and common invalid sentinels, such as extreme negative missing codes, are treated as missing values before quality control. This design prevents sentinel values from contaminating summary statistics and downstream models.

\subsection{City-level consolidation}

The first processing stage writes intermediate city-level caches in \texttt{CACHE\_DIR/<CITY>\_<UF>/combined\_part.csv}. The second stage repairs malformed CSV rows when possible, applies numeric coercion and quality control, and exports final compressed files named:

\begin{verbatim}
INMET_<CITY>_<UF>_<DATE_START>_<DATE_END>.csv.gz
\end{verbatim}

The city-level file organization is useful for batch processing, while the station fields preserve station identity for station-level filtering and modeling.

\section{Quality Control and Technical Validation}

\subsection{Two-stage quality-control strategy}

The quality-control protocol separates value invalidation from diagnostic flagging. Stage 1 applies broad physical bounds. Values outside the accepted interval are replaced by \texttt{NaN} and recorded with the \texttt{PHYSICAL\_BOUND} rule. Stage 2 applies temporal and cross-variable checks. These checks create flags but do not necessarily overwrite values, allowing users to decide whether to remove, retain, or reweight suspicious observations. This split between hard invalidation and traceable diagnostic screening is consistent with widely used practices in meteorological data quality assurance and station-dataset curation \cite{durre2010qa,dunn2012hadisd,dunn2016expanding_hadisd,lewis2021hourly_rain,nishimura2023sigma}.

For a station $s$, variable $j$, and time index $t$, the physical-bound rule can be summarized as:
\begin{equation}
x_{s,t}^{(j)} =
\begin{cases}
\mathrm{NaN}, & x_{s,t}^{(j)} < L_j \ \mathrm{or}\ x_{s,t}^{(j)} > U_j,\\
x_{s,t}^{(j)}, & \mathrm{otherwise},
\end{cases}
\end{equation}
where $L_j$ and $U_j$ are variable-specific lower and upper limits. When a bound is open-ended, only the available side is checked.

\subsection{Physical bounds}

Table~\ref{tab:physical_bounds} lists the Stage 1 limits implemented in the current pipeline.

\begin{table}[H]
\centering
\caption{Stage 1 physical bounds. Out-of-range values are converted to missing values and flagged as \texttt{PHYSICAL\_BOUND}.}
\label{tab:physical_bounds}
\scriptsize
\begin{tabular}{lll}
\toprule
Variable(s) & Accepted interval & Action\\
\midrule
\texttt{TEMP\_C}, \texttt{TEMP\_MAX\_C}, \texttt{TEMP\_MIN\_C} & $[-10, 45]$ $^\circ$C & set to \texttt{NaN} + flag\\
\texttt{RH\_\%}, \texttt{RH\_MAX\_\%}, \texttt{RH\_MIN\_\%} & $[0, 100]$ \% & set to \texttt{NaN} + flag\\
\texttt{PRECIPITATION\_MM} & $[0, +\infty)$ mm & set to \texttt{NaN} + flag\\
\texttt{WIND\_SPEED\_MPS}, \texttt{WIND\_GUST\_MPS} & $[0, 75]$ m s$^{-1}$ & set to \texttt{NaN} + flag\\
\texttt{WIND\_DIR\_DEG} & $[0, 360]$ degrees & set to \texttt{NaN} + flag\\
\texttt{SOLAR\_RADIATION\_KJ\_M2} & $[0, +\infty)$ kJ m$^{-2}$ & set to \texttt{NaN} + flag\\
\texttt{PRESSURE\_MB}, \texttt{PRESSURE\_MAX\_MB}, \texttt{PRESSURE\_MIN\_MB} & $[790, 1110]$ hPa & set to \texttt{NaN} + flag\\
\bottomrule
\end{tabular}
\end{table}

\subsection{Diagnostic consistency rules}

Table~\ref{tab:flags} summarizes the Stage 2 diagnostic rules. These rules are intentionally conservative: they highlight records that deserve attention, but they preserve the measurements unless the user chooses to invalidate them in a downstream filter.

\begin{longtable}{p{0.26\textwidth}p{0.27\textwidth}p{0.39\textwidth}}
\caption{Stage 2 diagnostic flags implemented in \datasetname.}\label{tab:flags}\\
\toprule
Flag & Variable & Meaning\\
\midrule
\endfirsthead
\toprule
Flag & Variable & Meaning\\
\midrule
\endhead
\texttt{PRECIP\_FROM\_PRESSURE\_LEAK} & \texttt{PRECIPITATION\_MM} & Precipitation value falls in a pressure-like range or is approximately equal to a pressure variable; value is replaced by \texttt{NaN}.\\
\texttt{TEMP\_SPIKE\_NO\_RAIN\_WIND} & \texttt{TEMP\_C} & Absolute hourly temperature change greater than 5 $^\circ$C without rain and without relevant wind change.\\
\texttt{RH\_JUMP\_NO\_RAIN\_WIND} & \texttt{RH\_\%} & Absolute hourly relative-humidity change of at least 50 percentage points without rain and without relevant wind change.\\
\texttt{RAIN\_GT\_200} & \texttt{PRECIPITATION\_MM} & Hourly precipitation greater than 200 mm; retained but flagged for verification.\\
\texttt{RAIN\_EVENT\_NON\_MONOTONIC} & \texttt{PRECIPITATION\_MM} & Decrease detected inside a positive-rain event block according to the implemented event heuristic.\\
\texttt{ISOLATED\_RAIN\_IN\_DRY\_PERIOD} & \texttt{PRECIPITATION\_MM} & Positive isolated precipitation between dry neighboring hours.\\
\texttt{WIND\_SPIKE\_0\_TO\_50} & \texttt{WIND\_SPEED\_MPS} & Jump from 0 to at least 50 m s$^{-1}$ in one hour.\\
\texttt{DIR\_CONSTANT\_2H} & \texttt{WIND\_DIR\_DEG} & Same wind direction for a three-point centered sequence, interpreted as at least two consecutive hours of constant direction.\\
\texttt{RADIATION\_AT\_NIGHT} & \texttt{SOLAR\_RADIATION\_KJ\_M2} & Positive global radiation during a UTC night window heuristic, defined as 22:00--09:00 UTC.\\
\texttt{PRESSURE\_JUMP\_GT5} & \texttt{PRESSURE\_MB} & Absolute hourly pressure change greater than 5 hPa.\\
\texttt{TEMP\_CONSISTENCY} & \texttt{TEMP\_C} & Instantaneous temperature violates the minimum--instantaneous--maximum relation.\\
\texttt{RH\_CONSISTENCY} & \texttt{RH\_\%} & Instantaneous humidity violates the minimum--instantaneous--maximum relation.\\
\texttt{PRESSURE\_CONSISTENCY} & \texttt{PRESSURE\_MB} & Instantaneous pressure violates the minimum--instantaneous--maximum relation.\\
\texttt{RAIN\_HOURLY\_MAX} & \texttt{PRECIPITATION\_MM} & Maximum hourly precipitation identified for each station series and saved as a diagnostic record.\\
\bottomrule
\end{longtable}

\subsection{Failure-hour summaries}

For each station and summarized variable, \datasetname{} computes the number of post-filter failed hours:
\begin{equation}
\mathrm{FAIL\_HOURS}_{s,j} = \sum_{t=1}^{T_s} \mathbb{I}\left[\mathrm{isna}\left(x_{s,t}^{(j)}\right)\right],
\end{equation}
where $T_s$ is the number of hourly records available for station $s$ and $\mathbb{I}$ is the indicator function. These summaries enable users to select stations according to completeness thresholds before training models or performing climatological analyses.

\begin{figure}[H]
\centering
\includegraphics[width=0.92\textwidth]{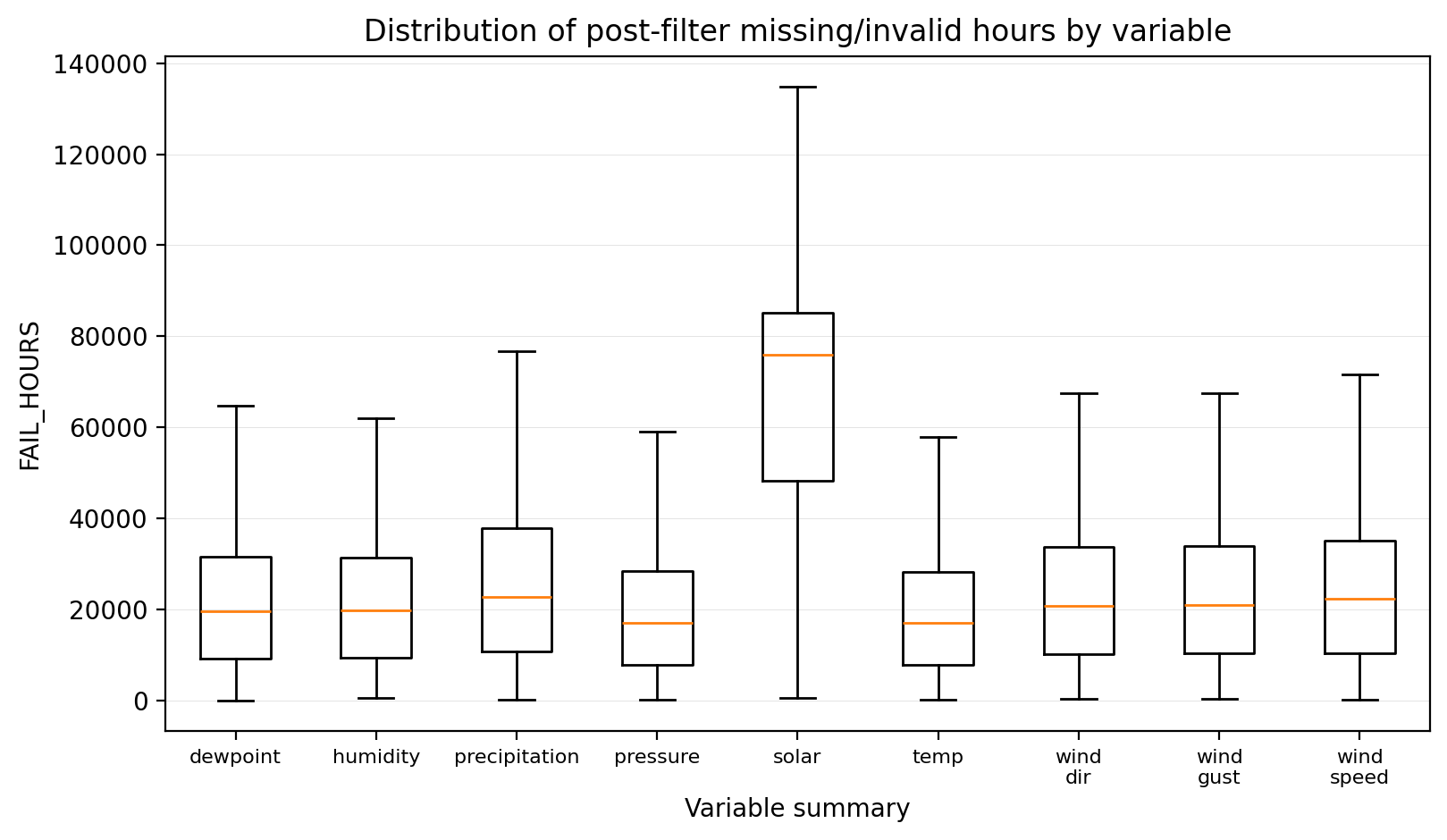}
\caption{Distribution of post-filter missing or invalid hours by variable summary. Outliers are hidden in the boxplot to improve readability.}
\label{fig:fail_hours}
\end{figure}

Figure~\ref{fig:fail_hours} shows the distribution of \texttt{FAIL\_HOURS} across the variable summaries in the current snapshot. Solar radiation presents the largest missing/invalid-hour counts, which is expected to require careful interpretation because radiation availability and night-time quality checks differ from scalar meteorological variables such as temperature and pressure.

\subsection{Geographic validation of station metadata}

\begin{figure}[H]
\centering
\includegraphics[width=0.72\textwidth]{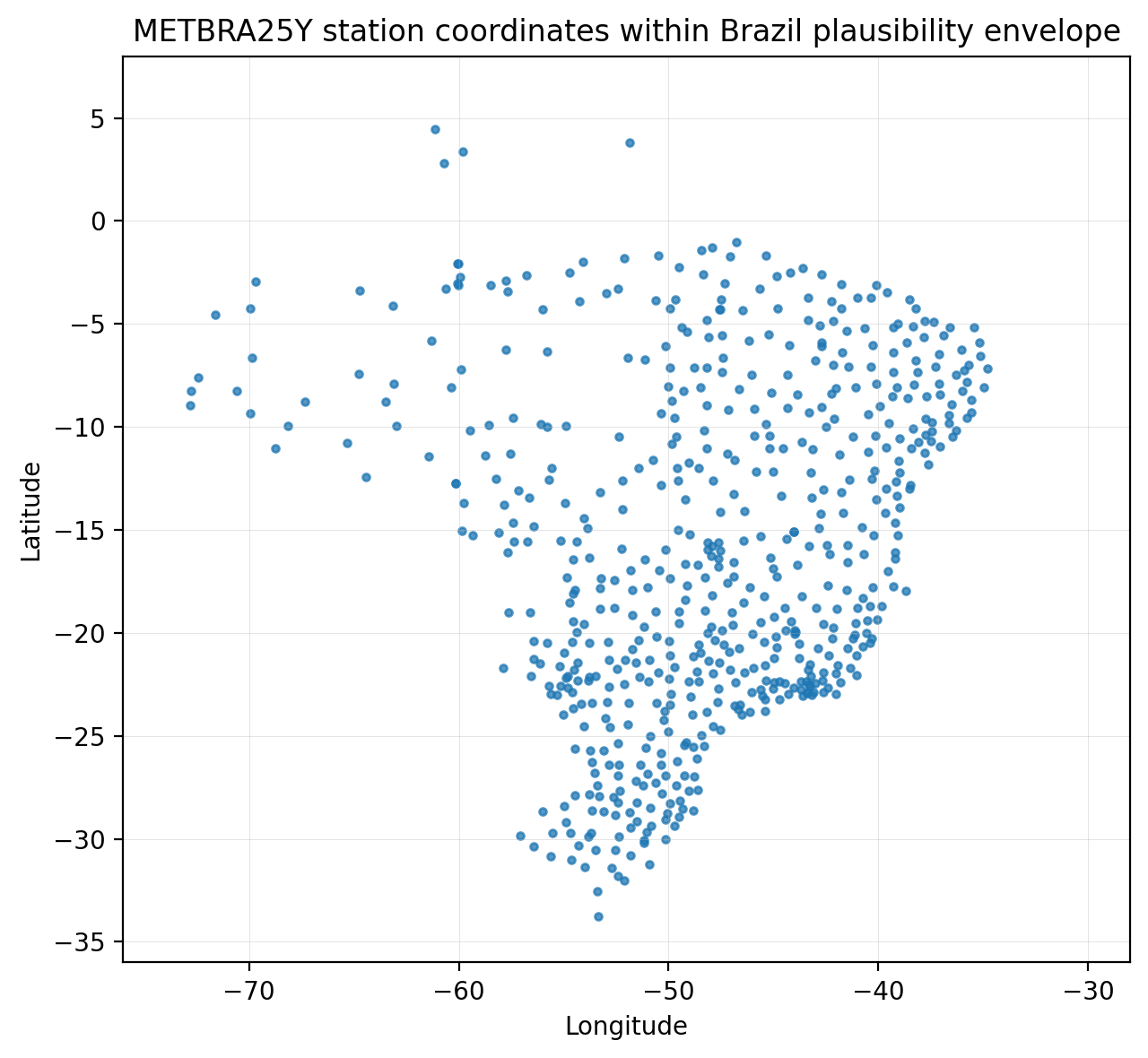}
\caption{Station coordinates retained after applying a broad Brazil plausibility envelope. This figure is a metadata validation diagnostic rather than a cartographic boundary map.}
\label{fig:station_map}
\end{figure}

A simple coordinate plausibility screen was applied to the station coordinates extracted from the summary files. Among the 616 unique station codes observed in the uploaded summaries, 605 stations fall within a broad Brazil envelope of latitude $[-40, 10]$ and longitude $[-80, -25]$. The remaining records should be manually checked because they appear to include malformed coordinates or locations outside the intended national surface-station scope. Figure~\ref{fig:station_map} displays the valid-coordinate subset.

\begin{figure}[H]
\centering
\includegraphics[width=0.76\textwidth]{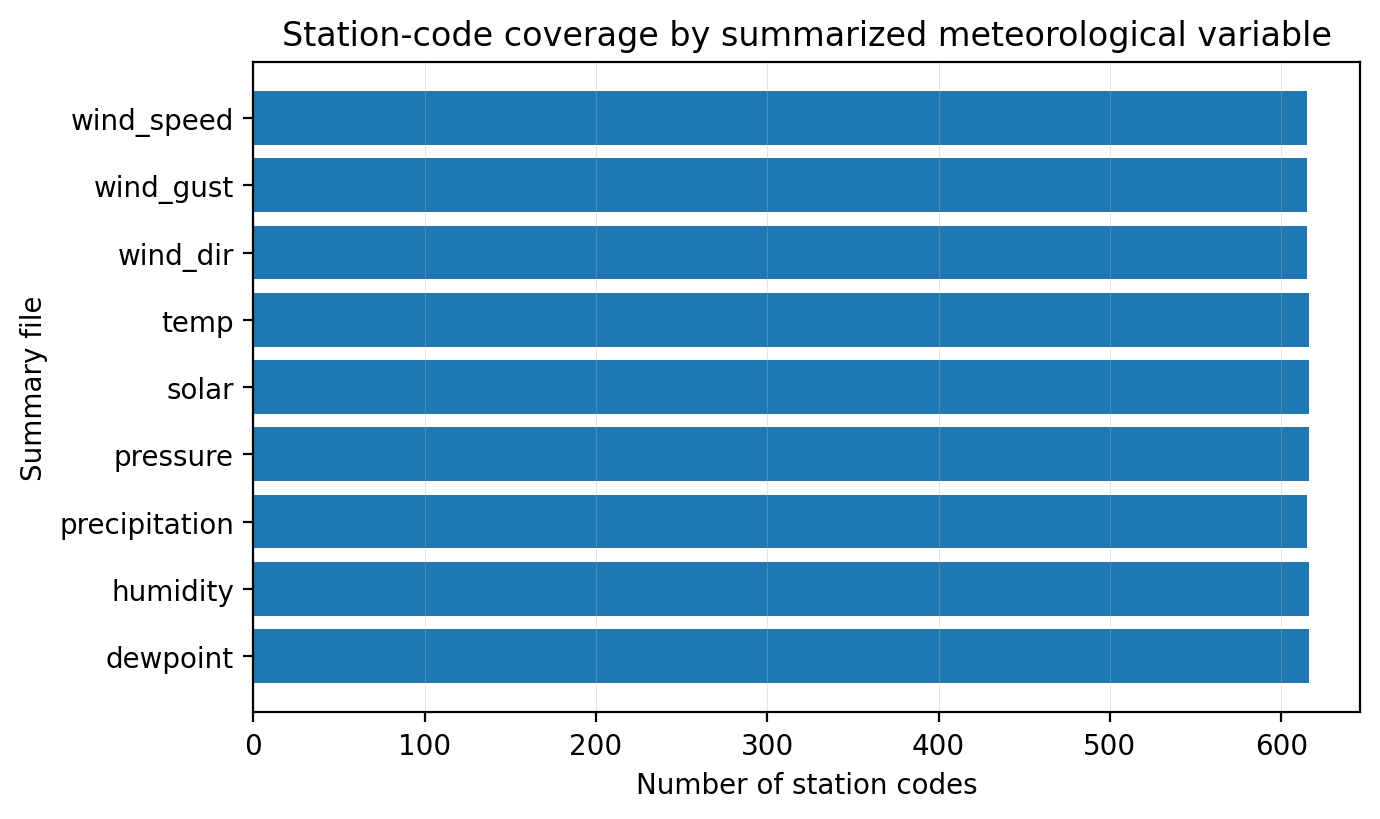}
\caption{Station-code coverage by variable-level summary file in the current release snapshot.}
\label{fig:station_counts}
\end{figure}

\begin{figure}[H]
\centering
\includegraphics[width=0.76\textwidth]{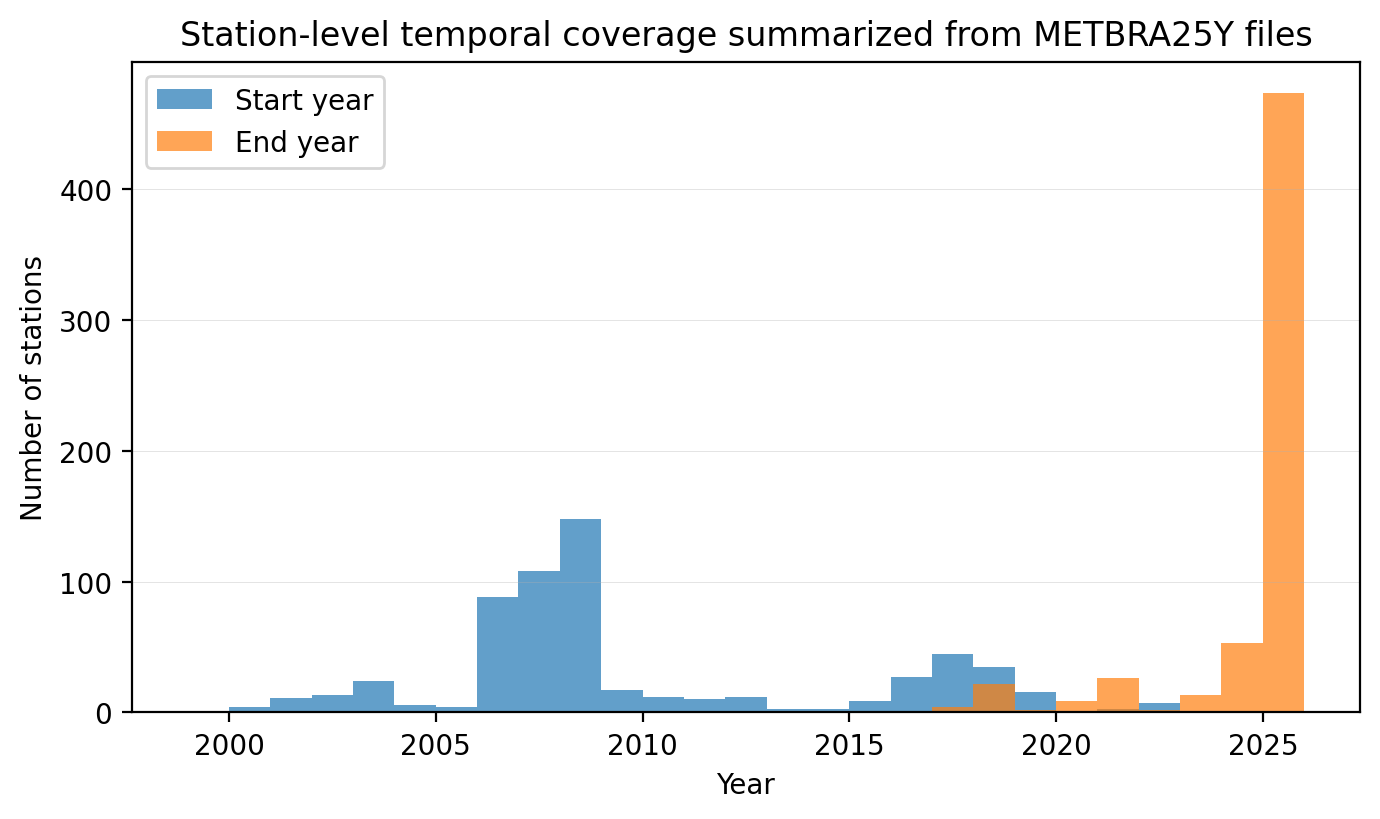}
\caption{Distribution of station-level start and end years computed from the summary files after coordinate plausibility filtering.}
\label{fig:coverage}
\end{figure}

\section{Data Records}

The release materials associated with \datasetversion{} document 20,067 processed raw INMET CSV files and 166,755,264 ingested observation rows across the 2000--2025 source archives. The current release snapshot also includes 616 unique station codes across the variable-level summaries, nine summary tables, and 1,231 per-station quality-control products, comprising 616 flag tables and 615 daily-precipitation files.

\subsection{Recommended package structure}

The complete \datasetname{} package released at Zenodo \cite{metbra25y_zenodo_2026} is organized according to the structure below. Temporary and cache directories are included for reproducibility but may be omitted from lightweight public releases if scripts can regenerate them.

\begin{verbatim}
METBRA25Y_release_v1.0.0/
|- README.md
|- LICENSE.md
|- CHANGELOG.md
|- CITATION.cff
|- SHA256SUMS.txt
|- code/
|  |- METBRA25Y.py
|  |- generatemap.py
|  \- requirements.txt
|- data/
|  |- city_files/
|  |  \- inmet_by_city_25y/
|  |     |- INMET_<CITY>_<UF>_2000-01-01_2025-12-31.csv.gz
|  |     |- manifest.csv
|  |     |- missing_summary.csv
|  |     \- missing_by_column.csv
|  \- manifest/
|- summaries/
|  |- precipitation_summary.csv
|  |- pressure_summary.csv
|  |- temp_summary.csv
|  |- dewpoint_summary.csv
|  |- humidity_summary.csv
|  |- wind_speed_summary.csv
|  |- wind_gust_summary.csv
|  |- wind_dir_summary.csv
|  \- solar_summary.csv
|- quality_flags/
|  |- <STATION_CODE>_flags.csv
|  \- <STATION_CODE>_precip_daily.csv
|- logs/
|  |- processed_files.csv
|  \- years_by_city.csv
|- dashboard/
|  \- METBRA_dashboard.html
\end{verbatim}

\subsection{Final city files}

The final data records are compressed CSV files grouped by city and state. Each row corresponds to an hourly observation and includes both meteorological variables and station metadata. The core columns include:

\begin{verbatim}
DATAHORA, STATION_CODE, STATION_NAME, LAT, LON,
PRECIPITATION_MM, PRESSURE_MB, PRESSURE_MAX_MB, PRESSURE_MIN_MB,
SOLAR_RADIATION_KJ_M2, TEMP_C, DEWPOINT_C, TEMP_MAX_C, TEMP_MIN_C,
RH_%, RH_MAX_%, RH_MIN_%,
WIND_DIR_DEG, WIND_GUST_MPS, WIND_SPEED_MPS
\end{verbatim}

\subsection{Manifest}

The manifest, \texttt{data/city\_files/inmet\_by\_city\_25y/manifest.csv}, is designed as the primary entry point for station discovery. It links station codes and metadata to the city-level file in which the station appears. Recommended manifest fields are:
\begin{verbatim}
city, uf, STATION_CODE, STATION_NAME, LAT, LON,
start_year, end_year, rows, file
\end{verbatim}

\subsection{Quality flags}

Per-station flags are saved as:
\begin{verbatim}
quality_flags/<STATION_CODE>_flags.csv
\end{verbatim}
with the schema:
\begin{verbatim}
STATION_CODE, DATAHORA, PARAM, RULE, VALUE, DETAIL
\end{verbatim}

This table is intentionally long-form so that users can filter by rule, variable, station, or time interval. For precipitation, the pipeline also saves daily aggregates:
\begin{verbatim}
quality_flags/<STATION_CODE>_precip_daily.csv
\end{verbatim}
with the columns \texttt{RAIN\_DAILY\_SUM} and \texttt{RAIN\_DAILY\_MEAN}.

\subsection{Variable-level summaries}

The \texttt{summaries/} directory provides compact station-level completeness summaries for the main meteorological variables. These files are appropriate for preselecting stations before loading the full hourly archive. For example, a user can restrict an analysis to stations with \texttt{FAIL\_HOURS} below a chosen threshold for temperature and precipitation.

\subsection{Interactive dashboard}

The script \texttt{generatemap.py} creates an offline HTML dashboard that displays station locations and selected meteorological series. In the current implementation, the dashboard reads the manifest, filters stations by plausible coordinates, samples stations when necessary to keep the HTML file lightweight, and plots daily precipitation, daily mean temperature, daily mean relative humidity, daily wind speed, and daily mean pressure for a recent multi-year window.

\section{Usage Notes}

\subsection{Loading the final records}

The final CSV.GZ files can be loaded directly with \textit{pandas}. Users should parse \texttt{DATAHORA} as a datetime column and explicitly document whether the resulting timestamps are interpreted as UTC-naive or localized.

\begin{verbatim}
import pandas as pd

path = "data/city_files/inmet_by_city_25y/INMET_<CITY>_<UF>_2000-01-01_2025-12-31.csv.gz"
df = pd.read_csv(path, compression="infer", low_memory=False)
df["DATAHORA"] = pd.to_datetime(df["DATAHORA"], errors="coerce")
\end{verbatim}

\subsection{Recommended filtering strategy}

For predictive modeling, users should avoid treating all stations and variables as equally complete. A recommended workflow is:
\begin{enumerate}
    \item read the relevant \texttt{summaries/*\_summary.csv} files;
    \item filter stations by temporal coverage and acceptable \texttt{FAIL\_HOURS};
    \item inspect \texttt{flags/<STATION\_CODE>\_flags.csv} for variables central to the analysis;
    \item decide whether Stage 2 flags should be removed, retained, winsorized, interpolated, or modeled with missingness indicators;
    \item report all thresholds and post-processing choices.
\end{enumerate}

\subsection{Examples of suitable applications}

\datasetname{} is suitable for:
\begin{itemize}
    \item rainfall forecasting and extreme-precipitation screening;
    \item hydrological and urban-flood susceptibility modeling;
    \item station-level climatological summaries;
    \item environmental-health studies requiring meteorological covariates;
    \item agricultural and renewable-energy analyses;
    \item benchmarking time-series imputation and forecasting models;
    \item spatiotemporal machine-learning studies using Brazilian meteorological stations.
\end{itemize}

These use cases are consistent with the types of downstream applications that have motivated both Brazilian meteorological products and broader station-based QC datasets, including gridded weather reconstruction, hydrometeorological modeling, and gap-filling workflows \cite{xavier2016brazil_gridded,xavier2022brdwgd,costa2021gapfill_qc}.

\section{Limitations and Considerations for Use}

\subsection{Source dependence}

\datasetname{} is a derived processing layer over public INMET records. Any changes, corrections, or gaps in the upstream INMET source will propagate to the derived archive unless explicitly documented and versioned. Users requiring legal or operational authority should cite and consult the original INMET records.

\subsection{Station-specific coverage}

The label ``25Y'' reflects the configured multi-year period, not continuous availability for every station. Many stations have shorter coverage windows, and each station-variable pair may present different missingness.

\subsection{Quality-control thresholds}

The implemented physical bounds and diagnostic rules are pragmatic and transparent, but they are not a substitute for expert station-by-station quality assessment. Extreme events may be flagged for verification even when physically real. Conversely, plausible but erroneous values may pass broad range checks.

\subsection{Solar radiation and night-time heuristics}

The current radiation-at-night rule uses a fixed UTC window rather than station-specific solar geometry. This heuristic is useful for screening but may overflag or underflag observations near daylight-transition periods. A future release should consider astronomical sunrise/sunset calculations using station latitude, longitude, and date.

\subsection{Coordinate metadata}

A small subset of station records in the current \datasetversion{} snapshot exhibits implausible coordinates, likely due to malformed decimal parsing or source-metadata inconsistencies. These records were retained in the release for transparency, and users are advised to apply coordinate plausibility filters before geospatial analyses.

\subsection{City-level grouping}

Although the final files are grouped by city, multiple stations can exist within a city or across nearby locations. Users should avoid assuming that a city-level file is equivalent to a spatially homogeneous city climatology. Station-level filtering by \texttt{STATION\_CODE} is recommended.

\section{Conclusion}

\datasetname{} provides a reproducible and harmonized archive of Brazilian hourly surface meteorological observations derived from INMET historical records. By combining canonical variable names, station metadata, compressed city-level files, explicit quality flags, daily precipitation aggregates, missing-data audits, and variable-level completeness summaries, the dataset reduces the preprocessing burden for researchers and practitioners working with Brazilian meteorological time series. The archive is particularly useful for environmental modeling, rainfall forecasting, hydrological risk analysis, and machine-learning experiments that require transparent handling of missing and suspicious observations.

Version \datasetversion{} is released with a Zenodo DOI, explicit licensing, citation metadata, checksum manifests, and a documented changelog. Future versions should continue to improve metadata consistency, especially for the small subset of implausible coordinates identified during technical validation, while preserving the transparent and versioned release model established in this first public release.

\section*{Funding}

This work received no specific external funding.

\section*{Acknowledgements}

The authors acknowledge the Instituto Nacional de Meteorologia (INMET) for making public meteorological observations available through its historical-data services and station catalog. The authors also acknowledge the institutional support of Sao Paulo State University (UNESP) and RECOD.ai, University of Campinas (UNICAMP), which helped make the curation and release of \datasetname{} possible.

\section*{Author contributions}

Matheus Lima Castro: conceptualization, data curation, software, methodology, formal analysis, visualization, writing--original draft. William Dantas Vichete: validation, data review, writing--review and editing. Leopoldo Lusquino Filho: supervision, conceptualization, methodology, writing--review and editing.

\section*{Competing interests}

The authors declare no competing interests.

\end{document}